\documentclass[pdflatex,sn-mathphys-num]{sn-jnl}% Math and Physical Sciences Numbered Reference Style
\usepackage{graphicx}%
\usepackage{multirow}%
\usepackage{amsmath,amssymb,amsfonts}%
\usepackage{amsthm}%
\usepackage{mathrsfs}%
\usepackage[title]{appendix}%
\usepackage{xcolor}%
\usepackage{textcomp}%
\usepackage{manyfoot}%
\usepackage{booktabs}%
\usepackage{listings}%

\usepackage{comment}
\usepackage{array}
\usepackage[caption=false,font=normalsize,labelfont=sf,textfont=sf]{subfig}
\usepackage{textcomp}
\usepackage{stfloats}
\usepackage{url}
\usepackage{verbatim}
\usepackage{graphicx}
\usepackage{color}
\usepackage[linesnumbered,ruled,vlined]{algorithm2e}

\theoremstyle{thmstyleone}%
\theoremstyle{thmstyletwo}%

\theoremstyle{thmstylethree}%

\begin{document}

\title[Article Title]{Vision-Based Embedded System for Noncontact Monitoring of Preterm Infant Behavior in Low-Resource Care Settings}


\author*[1]{\fnm{Stanley} \sur{Mugisha [0000-0002-0046-6850]}}\email{smugisha@sun.ac.ug} 

\author[1,3]{\fnm{Rashid} \sur{Kisitu}}\email{kisiturashid01@gmail.com }

\author[1]{\fnm{Francis} \sur{Komakech}}\email{fkomakech@sun.ac.ug}

\author[2]{\fnm{Excellence} \sur{Favor}}\email{efavor@sun.ac.ug}

\affil*[1]{\orgdiv{Electronics and Computer Engineering}, \orgname{Soroti University}, \orgaddress{\street{Arapai}, \city{Soroti},  \country{Uganda}, }}

\affil[2]{\orgdiv{Electrical and Energy Engineering}, \orgname{Soroti University}, \orgaddress{\street{Arapai}, \city{Soroti City}, \country{Uganda}}}

\affil[3]{\orgname{Batmaak Engineering}, \orgaddress{\street{Katwe}, \city{Kampala},  \country{Uganda}}}


\abstract{Preterm birth remains a leading cause of neonatal mortality, disproportionately affecting low-resource settings with limited access to advanced neonatal intensive care units (NICUs).
Continuous monitoring of infant behavior, such as sleep/awake states and crying episodes, is critical but relies on manual observation or invasive sensors, which are prone to error, impractical, and can cause skin damage. 
This paper presents a novel, noninvasive, and automated vision-based framework to address this gap. We introduce an embedded monitoring system that utilizes a quantized MobileNet model deployed on a Raspberry Pi for real-time behavioral state detection. When trained and evaluated on public neonatal image datasets, our system achieves state-of-the-art accuracy (91.8\% for sleep detection and 97.7\% for crying/normal classification) while maintaining computational efficiency suitable for edge deployment. 
Through comparative benchmarking, we provide a critical analysis of the trade-offs between model size, inference latency, and diagnostic accuracy. Our findings demonstrate that while larger architectures (e.g., ResNet152, VGG19) offer marginal gains in accuracy, their computational cost is prohibitive for real-time edge use. The proposed framework integrates three key innovations: model quantization for memory-efficient inference (68\% reduction in size), Raspberry Pi-optimized vision pipelines, and secure IoT communication for clinical alerts. This work conclusively shows that lightweight, optimized models such as the MobileNet offer the most viable foundation for scalable, low-cost, and clinically actionable NICU monitoring systems, paving the way for improved preterm care in resource-constrained environments.}

\keywords{Neonatal Monitoring, Embedded AI, Computer Vision, MobileNet, Edge Computing, Low-Resource Healthcare, NICU, Preterm Infant, Model Quantization, Raspberry Pi, IoT-enabled, Machine Learning}

\maketitle
\section{Introduction}\label{sec1}
Preterm birth, defined as delivery before 37 weeks of gestation, remains a critical global health challenge and accounts for approximately 35\% of neonatal deaths worldwide \cite{who2025preterm}. This burden is most severe in low-resource settings, where scarce access to advanced NICUs and a shortage of trained staff exacerbate the risks for vulnerable infants \cite{Lawn2014EveryNewborn,  frontiers2022pretermburden}. The survival and long-term health of preterm infants depends on the continuous, precise monitoring of behavioral states such as sleep patterns, awake activity, crying and distress signals \cite{who2025preterm}. 

Crying and sleep offer communication cues for infants and therefore monitoring them among infants is crucial because; Infant crying is a primary communication tool that signals a baby's needs, such as hunger, discomfort, or pain \cite{Hammoud2024}. Correctly interpreting these cues is essential for timely care. For a baby who can't speak, a prolonged, high-pitched cry can be the first sign of a serious condition like sepsis or respiratory distress. Similarly, sleep is a vital indicator of an infant's neurological and physical development. Abnormal sleep patterns can signal potential issues, and analysis is key to early diagnosis of pediatric neurological disorders \cite{leo2022}. The ability to accurately and continuously monitor these two key behaviors provides a powerful, non-invasive way to assess infant well-being.

Current NICU practices face significant limitations in monitoring these critical behaviors. Manual observation by clinicians is labor-intensive, subjective, and prone to delayed recognition of critical events \cite{memon2020}. Invasive wearable sensors, while providing objective data, can cause skin irritation, restrict mobility, and hinder essential care practices like kangaroo mother care \cite{pubmedcentral2023wireless}.
The advent of computer vision and deep learning presents contactless monitoring which is a promising alternative to traditional methods. Techniques like facial expression analysis \cite{Fotiadou2014,liang2022,Gupta2021_2,gupta2021} and remote photoplethysmography (rPPG) \cite{Poh2010} can extract behavioral and physiological cues from video feeds. For instance, convolutional neural networks (CNNs) have been used to detect infant crying through facial landmark tracking \cite{huang2019}, and video analytics have shown promise in the early diagnosis of neurological disorders \cite{leo2022}.

Despite these innovations, a significant deployment gap exists as most state-of-the-art models such as ResNet and Inception, prioritize accuracy over computational efficiency, and hence unsuitable for real-time inference on edge devices with limited processing power and memory \cite{tan2019, arxiv2024edgelearning}. 

Thus this study closes the gap between AI innovation and clinical practicality by introducing a vision-based IoT-enabled framework for real-time neonatal behavioral monitoring. We present an optimized and quantized MobileNetV3 model deployed on a Raspberry Pi 5 to create an end-to-end system that operates on  low-cost, off-the-shelf hardware. This system captures infant behavior via a low-power camera, processes video streams locally on the edge device, and transmits actionable insights to healthcare providers via a secure mobile interface. Our approach eliminates dependency on centralized servers, mitigates privacy risks, and ensures operation during network outages, which is critical for NICUs in regions with unreliable internet infrastructure . Our study makes three key contributions:
\begin{enumerate}
    \item A Novel Edge-Optimized Architecture. We present the integration of a quantized MobileNet model for preterm infant monitoring. This model achieves high accuracy and low latency for real-time inference on embedded hardware.
    \item A Comprehensive Embedded Framework: We design and deploy a full-stack, end-to-end system that integrates optimized computer vision, on-device inference, and secure IoT communication, thus provide a practical blueprint for clinical deployment in low-resource settings.
    \item In-depth performance analysis: We provide a rigorous benchmark studies that quantify the critical trade-offs between model accuracy, latency, and the memory footprint, for embedded NICU monitoring and offer valuable insights for future research and development.
\end{enumerate}

This work paves the way for scalable, cost-effective solutions to reduce preterm mortality in underserved regions. The following sections detail the methodology, experimental results, and deployment pipeline, concluding with insights into future directions for vision-based neonatal care.

\section{Related Work}
\label{sec:related-work}

The shift towards autonomous infant health monitoring using artificial intelligence, particularly deep learning, represents a significant advancement. This section provides a critical review of existing literature, highlighting the methodologies, performance, and key limitations of current approaches in infant cry classification and sleep stage detection. We also examine the emerging field of image-based monitoring, identifying a crucial gap in the literature that our work addresses. Our goal is to demonstrate that while high-accuracy models exist, they are often impractical for real-world deployment on low-cost hardware in resource-constrained environments.

\subsection{Audio and physiological signal-based approaches}
Current literature on infant health monitoring predominantly uses audio or contact-based physiological sensors. While these approaches have achieved high accuracy, they suffer from fundamental limitations in terms of real-world applicability.

\subsubsection{Infant cry classification}
Automated infant cry classification aims to identify a baby's needs (e.g., hunger, pain) or pathologies (e.g., respiratory distress) from audio signals. Research has progressed from traditional machine learning models using handcrafted features like MFCCs \cite{chang2021,Hammoud2024} to complex deep learning architectures \cite{Ashwini2021,Li2024,Zayed2023}. While some of these models claim accuracies exceeding 95\%, a significant gap exists between their laboratory performance and practical deployment.
The field overwhelmingly prioritizes accuracy, often employing computationally intensive models like transformers and large deep neural networks (DNNs) \cite{Li2024,Zayed2023}. These models are often trained on small, clean datasets, and their performance in noisy, real-world hospital environments is largely unproven. More importantly, the literature almost universally ignores the metrics essential for embedded deployment: inference latency, memory footprint, and power consumption. Furthermore, these methods require a reliable audio capture infrastructure and can be susceptible to environmental noise, posing significant challenges in a real-world NICU. This lack of focus on computational efficiency and environmental robustness renders these models impractical for real-time monitoring on low-cost edge devices.

\subsubsection{Infant sleep stage classification}
Sleep stage classification is crucial for diagnosing pediatric sleep disorders. Traditional methods rely on polysomnography (PSG) data from contact-based sensors such as electroencephalography (EEG), electrooculography (EOG), and electromyography (EMG) \cite{phan2022,JIMENEZGARCIA2024}. Recent work has achieved high accuracy (up to 94.88\%) using sophisticated deep learning methods and multi-domain feature fusion on minimal EEG channels \cite{Irfan2024,Irfan2025}.
However, these methods are inherently contact-based and intrusive. The requirement for adhered sensors is fundamentally incompatible with the goal of non-invasive, continuous monitoring of preterm infants. These sensors can cause skin irritation, hinder essential care like kangaroo care, and require constant oversight from specialized staff. This makes them impractical for the long-term, scalable monitoring needed in low-resource settings, where non-invasive solutions are essential.

\subsection{Image-based monitoring}
The limitations of audio and contact-based methods have led to the exploration of non-contact, image-based approaches due to its potential to provide continuous monitoring without physical intrusion.
Early efforts used basic computer vision techniques for simple motion analysis. The introduction of deep learning, particularly Convolutional Neural Networks (CNNs), has enabled the analysis of video feeds to extract subtle physiological and behavioral cues \cite{medrxiv2025respiratory}. This has led to advancements in detecting infant cries from salient facial points \cite{Fotiadou2014,liang2022,Gupta2021_2,gupta2021} and assessing movements for diagnosing neurological disorders \cite{Gleason2024}. The application of lightweight CNNs, such as the MobileNetV3V2 architecture, has shown remarkable success in medical image classification tasks, such as classifying brain tumors from MRI scans with 99.16\% accuracy \cite{Adamu2024}. This demonstrates the potential of efficient architectures for complex visual analysis in medical contexts. 
Despite these innovations, the computational complexity of state-of-the-art deep learning models presents a significant hurdle. Architectures like ResNet and Inception, while highly accurate, demand substantial processing power and memory \cite{howard2019,arxiv2024edgelearning}. This makes them unsuitable for real-time inference on low-cost edge devices like the Raspberry Pi, which are common in affordable medical solutions \cite{howard2019}. This fundamental trade-off between model accuracy and computational efficiency is a critical, yet often unaddressed, consideration for practical deployment \cite{arxiv2025tradeoff}. Additionally, the majority of research in computer vision for infant monitoring has focused on algorithmic accuracy rather than computational efficiency and embedded deployment. As a result, prior work has largely been confined to powerful, server-grade hardware, creating a significant barrier to adoption in the low-resource settings that would benefit most from these technologies. 

The existing literature highlights a critical divide. While high-accuracy models exist for both audio and physiological signal analysis, they are either intrusive or computationally prohibitive for practical, real-world deployment in low-resource settings. Image-based approaches offer a promising noninvasive alternative, but the focus has remained on maximizing accuracy on powerful hardware, neglecting the need for efficiency on edge devices. Our work presents a novel approach that focuses on the holistic co-design of an efficient model architecture, aggressive optimization techniques, and deployment on a representative low-cost hardware platform. We use a lightweight MobileNetV3 architecture, combined with quantization, to develop a model that is accurate and computationally efficient. We validated our approach on a Raspberry Pi and thus provide a concrete pathway for real-world implementation, overcoming the key barriers identified in the literature and enabling the scalable, noninvasive monitoring of infants where it is needed most.

\section{Methodology}
\subsection{Research design}
The work adopted an experimental research design combined with embedded systems development techniques to create and validate a real-time behavioral monitoring system for NICUs. The approach consisted of two main stages: (1) the development and training of deep learning models for sleep/awake and crying/normal behavior classification, and (2) the integration of these models into an embedded system prototype for testing in a simulated NICU environment.

\subsection{Dataset acquisition and preprocessing}
\subsubsection{Dataset acquisition}
The works utilized two prerecorded datasets obtained from public repositories on the Roboflow platform and accessible at \cite{Rashid2025}. 
Two task-specific datasets were used: 
\begin{enumerate}
    \item  a sleep/awake dataset comprising 2,522 images partitioned into training/validation/test splits of 2,090/216/216 respectively;
    \item  a crying/normal dataset containing 4,099 images images partitioned into 3,144/784/171 for training/validation/test.
\end{enumerate}
Labels were assigned according to visible facial cues (eye closure, mouth openness, facial tension). 
The datasets were preannotated and reportedly had personally identifiable information removed by the dataset owners, adhering to privacy protocols. A robust augmentation pipeline was applied to improve generalization and simulate NICU conditions: random brightness/contrast adjustments ($\pm$30\%), Gaussian blur (kernel size up to 5x5), horizontal flipping, rotation ($\pm$15°), and zoom ($\pm$10%).

\subsection{Simulations}
The simulation involved feeding the prerecorded image datasets to the system to mimic receiving a live video feed. Real-time inference simulations were performed using prerecorded video streams at a frame rate of 25 FPS. OpenCV was used for basic image processing, such as resizing and normalization. Mediapipe detected facial landmark positions for sleep (eye closures) and crying (mouth openness) recognition. TensorFlow models classify individual frames to determine the behavioral state. The classification confidence thresholds were adjusted to optimize accuracy and reduce false positives and negatives. 

\subsection{Model architecture and training}
A MobileNetV3-Small backbone, pretrained on ImageNet, was used. The model incorporates native Squeeze-and-Excitation (SE) attention blocks for enhanced feature recalibration. SE block complexity was reduced by 30\% through channel pruning, preserving accuracy while lowering memory usage. The final classification head was replaced with a single dense layer with Softmax activation.

\subsubsection{Training protocol} Models were fine-tuned using TensorFlow 2.10 on an NVIDIA GTX 1080 GPU. The key hyperparameters included, the Adam optimizer (initial lr=0.001, reduced on plateau), categorical cross-entropy loss with label smoothing (factor=0.1) to mitigate class imbalance and improve calibration, a batch size of 32, and regularization via early stopping (patience=10 epochs) and dropout (20\%).

\subsection{Real-time behavioral processing pipeline}
The real-time behavioral processing pipeline in Figure \ref{fig:architecture} and algorithm  \ref{alg:recognition_pipeline} processes video frames in real-time to detect and classify neonatal behavioral states through three coordinated stages. 

\subsubsection{Stage 1. Face detection and preprocessing}
Input frames are first converted from the BGR to RGB color space to align with MediaPipe’s expected input format. A lightweight facial landmark detector localizes infant faces in the frame, generating bounding boxes with sub-millisecond latency. Valid detections meeting minimum confidence thresholds for facial feature visibility, are extracted and resized to 256×256 pixels, ensuring compatibility with the downstream classification model while preserving critical spatial details such as eye closure and mouth openness.

\subsubsection{Stage 2. Emotion prediction and temporal filtering} 
Valid facial regions are cropped, standardized and pixel values are normalized to a [0, 1] range for compatibility with the MobileNet classifier. A Mobilenet, processes each region of interest (ROI) to predict behavioral states such as sleep/awake or crying/normal states. 
Confidence scores were computed using Platt scaling to calibrate model outputs for clinical reliability. 
Predictions are further refined through temporal consistency checks, where a five-frame sliding window (200 ms at 25 FPS) smooths sporadic errors and triggers alerts only for sustained events, such as crying episodes exceeding 10 seconds.
Counters tracking behavioral duration (e.g., cumulative crying time) are updated only when stable states persist across consecutive frames.

\subsubsection{Stage 3. Annotation and asynchronous logging}
The original frame is annotated with bounding boxes, predicted labels, and confidence scores via OpenCV’s drawing utilities. To balance real-time responsiveness with cloud synchronization overhead, Firebase updates are batched and scheduled at 15second intervals. This ensures that critical events (e.g., prolonged crying) are logged immediately, while routine state transitions are aggregated to minimize network usage. Pending tasks—including frame rendering on the TFT display and diagnostic metric updates—are executed in a dedicated thread pool to prevent blocking the primary inference loop.

\begin{algorithm}
\DontPrintSemicolon
\SetAlgoLined
\KwData{Raw video frame}
\KwResult{Emotion prediction and annotation pipeline}
\DontPrintSemicolon
\SetKwProg{Procedure}{Procedure}{}{}
\Procedure{PROCESS\_MODEL\_FRAME(frame)}{
    Convert frame to RGB\;
    Detect faces using MediaPipe\;
    \ForEach{detected face}{
        Extract face bounding box\;
        \If{face is valid}{
            Preprocess face (resize to 256$\times$256)\;
            Predict emotion using current model\;
            Calculate confidence scores\;
            Update counters and timers based on prediction\;
            Annotate frame with prediction results\;
            
            \If{time\_to\_update\_firebase}{
                Prepare data packet\;
                Schedule Firebase update\;
            }
        }
    }
    Run pending scheduled tasks\;
    Update display metrics\;
    \Return{annotated\_frame}\;
}
\caption{Facial emotion prediction and annotation pipeline}
\label{alg:recognition_pipeline}
\end{algorithm}

\subsection{Edge deployment and optimization}
The transition from simulation to real-world NICU deployment requires meticulous hardware-software codesign.
\subsubsection{Hardware configuration}
\begin{itemize}
    \item Edge device. Raspberry Pi 5 (8 GB RAM, 64-bit quad-core Cortex-A76 CPU) running Raspberry Pi OS Lite.
    \item Imaging. Raspberry Pi Camera Module 3 (12 MP Sony IMX708 sensor) configured at 25 FPS (640×480 resolution) with HDR enabled for low-light resilience.
    \end{itemize}

\subsubsection{Software optimization}

\begin{enumerate}
    \item Model quantization: Trained models were converted to TensorFlow Lite format using FP16 quantization, reducing size by ~60\% and latency by 22\%.
    \item Multithreading: OpenCV-based preprocessing and inference tasks were parallelized across the Pi’s CPU cores, achieving a 1.8x speedup.
    \item Security: MQTT communication with TLS v1.2 encryption ensures secure data transmission to the Firebase cloud backend.
\end{enumerate} 

\subsection{Holistic evaluation}
Efficacy was assessed through dual lenses. clinical accuracy and embedded performance. 
\subsubsection{Clinical Metrics} These included 
\begin{itemize}
    \item Precision/Recall Trade-off. The precision (92.1\% for sleep detection), recall (91.8\%), and F1-score (91.9\%) were prioritized to minimize both false alarms and missed critical events.
    \item Temporal consistency. A 5-frame moving average filter suppressed transient misclassifications (e.g., brief eye closures mislabelled as sleep).
    \item AUC-ROC analysis.   A value of 0.98 was achieved for crying detection, demonstrating robust separability between distress and normal states.
\end{itemize}

\subsubsection{Embedded performance} The metrics included:
\begin{itemize}
    \item Latency.The end-to-end processing time per frame averaged 0.87 seconds, meeting the $ \leq$1 sec target for real-time alerts.
    \item Memory Footprint. The peak RAM usage remained below 1.2 GB during stress tests, ensuring compatibility with low-cost edge devices.
\end{itemize}

\subsection{Architectural setup}
\begin{figure}
    \centering
    \includegraphics[width=1.0\linewidth]{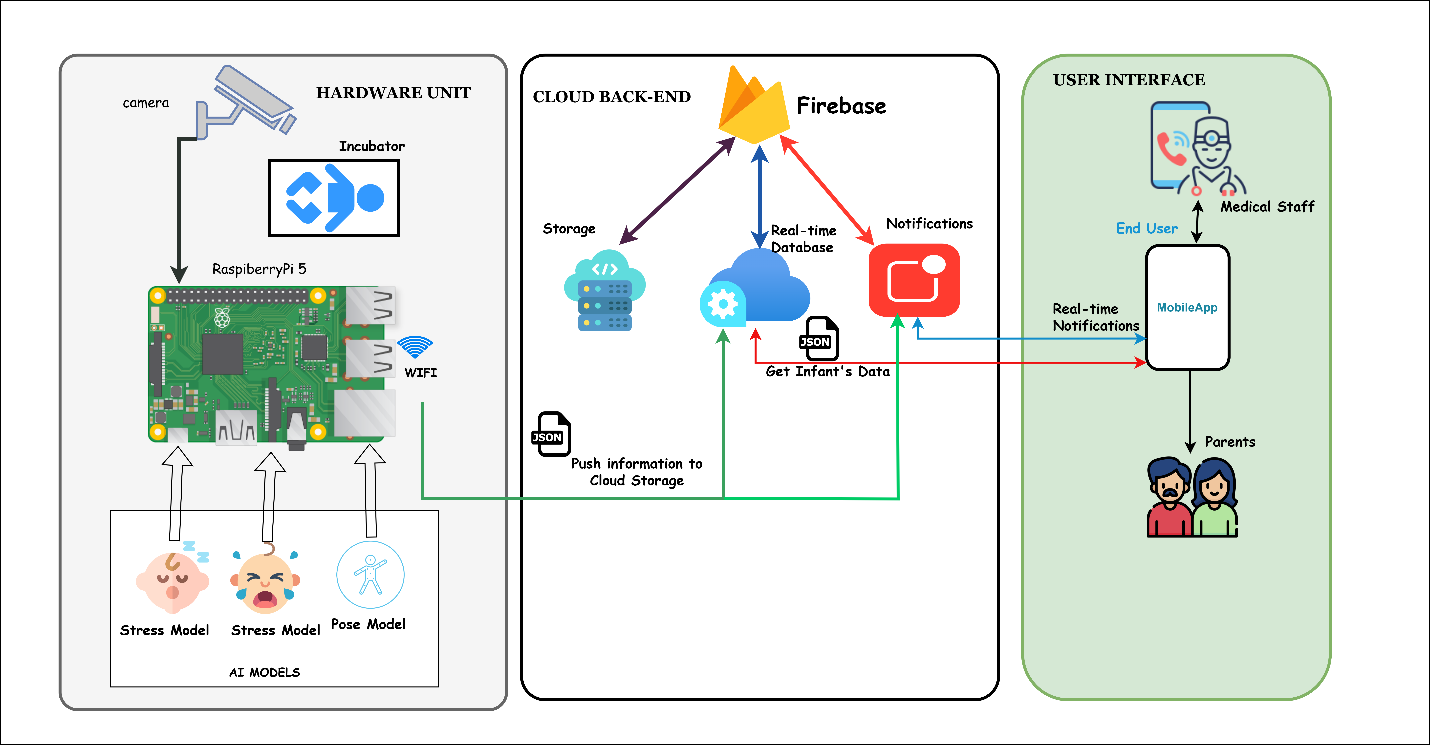}
    \caption{The architecture of the Neonatal behavioural monitoring system for preterm babies in NICUs.}
    \label{fig:architecture}
\end{figure}

The overall architecture is a three-tiered structure as shown in Figure \ref{fig:architecture}. The initial processing occurs locally on the Raspberry Pi, reducing latency for real-time analysis. The processed data is then sent to a cloud backend (Firebase) for centralized storage, synchronization across devices, and triggering alerts to the user interface (mobile app).

The architecture consists of the following main components:
\subsubsection{Hardware Unit (Edge device)}
Designed to be placed near the infant, likely within or near an incubator, is a pi camera connected to a Raspberry Pi 5. and connected to the internet via WiFi. 
 \begin{figure}
    \centering
    \includegraphics[width=0.8\linewidth]{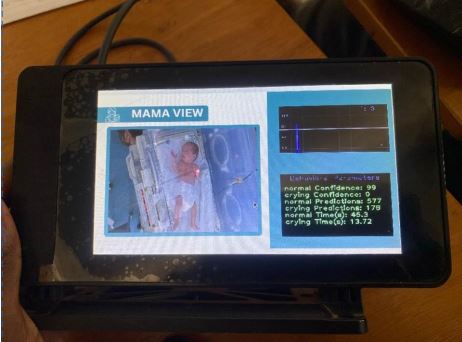}
    \caption{The models deployed on the raspberry pi for realtime inference and benchmarking.}
    \label{fig:rasp_pi}
\end{figure}
\subsubsection{Cloud back end}
 This is the central server-side component, which uses Firebase which provides a real-time database for storing and synchronizing data.

\subsubsection{User interface}
\begin{figure}
    \centering
    \includegraphics[width=0.4\linewidth]{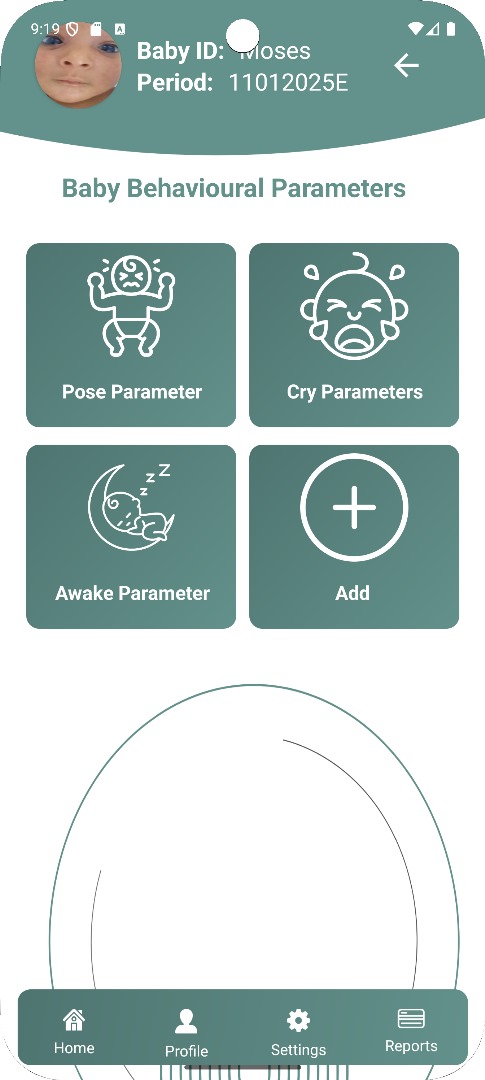}
    \caption{A screenshot of the mobile app interface for baby monitoring showing the three parameters for pose estimation, cry detection and sleep monitoring.}
    \label{fig:mob_app}
\end{figure}

The end user interface runs on a handheld device such as a mobile phone with an android mobile application as shown in Fig. \ref{fig:mob_app} or a raspberry pi as shown in Fig. \ref{fig:rasp_pi}. It allows for remote monitoring, user interaction (medical staff and parents) and monitoring and also receives real-time notifications from the cloud backend. 

\section{Results and discussions}
\subsection{Clinical accuracy benchmarking}
\subsubsection{Comparative analysis on different classification tasks}
The models were evaluated on standalone classification tasks. The results (Tables \ref{tab:awake_model_performance} - \ref{tab:crying_model_performance}) reveal the fundamental accuracy-efficiency trade-off.

\begin{table}[ht]
\centering
\caption{Performance metrics on the Awake/Sleep classification tasks.}
\label{tab:awake_model_performance}
\begin{tabular}{lcccccc}
\toprule
\textbf{Model Name} & \textbf{Acc (\%)} & \textbf{Precision} & \textbf{F1} & \textbf{AUC} & \textbf{Recall} & \textbf{Size (MB)} \\
\midrule
MobileNet (ours)      & 91.81 & 0.92 & 0.92 & 0.98 & 0.92 & 21.2 \\
LeNet          & 83.63 & 0.84 & 0.84 & 0.92 & 0.84 & 4.6 \\
EfficientNetB0 & 59.65 & 0.64 & 0.59 & 0.68 & 0.60 & 52.6 \\
ResNet152V2    & 88.89 & 0.89 & 0.89 & 0.96 & 0.89 & 236.0 \\
Inception      & 94.15 & 0.94 & 0.94 & 0.99 & 0.94 & 254.0 \\
\bottomrule
\end{tabular}
\end{table}

\begin{table}[ht]
\centering
\caption{Performance Metrics on Crying/Normal classification tasks}
\label{tab:crying_model_performance}
\begin{tabular}{lccccc}
\toprule
\textbf{Model Name} & \textbf{Acc (\%)} & \textbf{Precision} & \textbf{F1} & \textbf{AUC} & \textbf{Recall}  \\
\midrule
MobileNet      & 97.69 & 0.98 & 0.98 & 1.00 & 0.98  \\
LeNet          & 90.74 & 0.91 & 0.91 & 0.97 & 0.91  \\
EfficientNetB0 & 78.24 & 0.83 & 0.77 & 0.93 & 0.78  \\
ResNet152V2    & 90.74 & 0.91 & 0.91 & 0.97 & 0.91  \\
Inception      & 94.44 & 0.94 & 0.94 & 0.99 & 0.94 \\
\bottomrule
\end{tabular}
\end{table}

\begin{figure}
    \centering
    \includegraphics[width=1.0\linewidth]{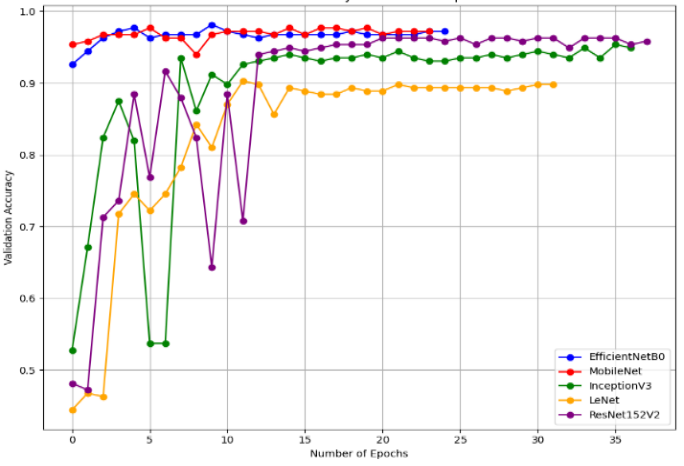}
    \caption{Training Results from the sleep/awake classification task}
    \label{fig:sleep_awakeGraph}
\end{figure}

\begin{figure}
    \centering
    \includegraphics[width=1.0\linewidth]{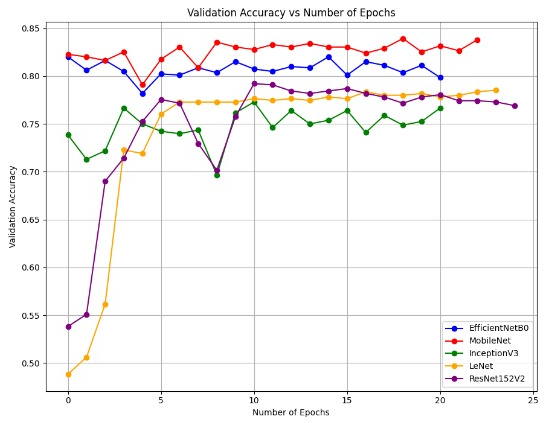}
    \caption{Training Results from the crying/normal classification task}
    \label{fig:emotionsGraph}
\end{figure}

\begin{figure}
    \centering
    \includegraphics[width=1.0\linewidth]{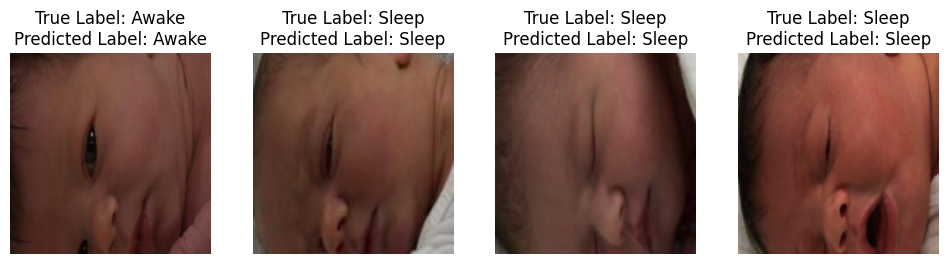}
    \caption{Visual mobilenet results from the awake sleep dataset}
    \label{fig:Awakesleep}
\end{figure}

\begin{figure}
    \centering
    \includegraphics[width=1.0\linewidth]{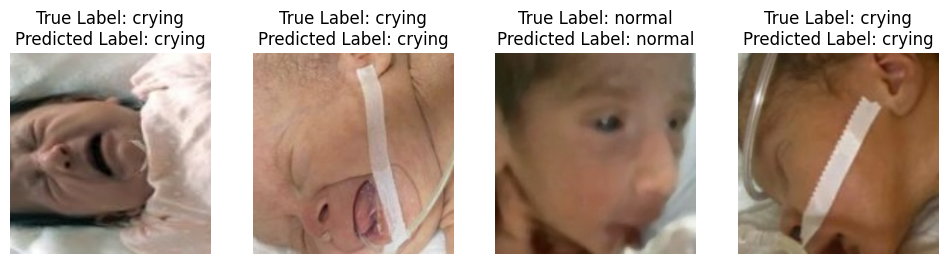}
    \caption{Visual mobilenet model results from the crying normal dataset}
    \label{fig:cryingnormal}
\end{figure}

The evaluation of the awake/sleep classification task in Table \ref{tab:awake_model_performance} reveals critical insights into the trade-offs between model complexity and clinical effectiveness. 
Inception emerged as the top performer with 94.15\% accuracy and a near perfect AUC of 0.99. The model shows its superior ability to detect subtle behavioral cues such as partial eye closures. However, its impracticality for edge deployment because of its large size of 254 MB offset its technical superiority.

Our MobileNet was slightly less accurate at 91.81\%, demonstrating an optimal compromise between performance and efficiency, achieving near-state-of-the-art accuracy with a 12x smaller model, demonstrating its suitability for the target task. Its lightweight architecture (21.2 MB), which leverages depthwise separable convolutions and squeeze-excitation blocks, proved adept at extracting eye region dynamics essential for sleep detection. Fig. \ref{fig:Awakesleep} shows sample classification results based on the test set.
Although ResNet152V2 achieved moderate accuracy (88.89\%), showed residual connections’ effectiveness in mitigating gradient vanishing for deeper networks. However, it suffered from over parameterization (236 MB), underscoring the challenges of deploying deep residual networks in resource limited NICUs. The LeNet’s shallow design limited its effectiveness, particularly in distinguishing transient eye movements from sustained sleep states. This model demonstrates the limitations of shallow architectures in neonatal monitoring. While computationally frugal, its inability to model hierarchical spatial relationships led to frequent misclassifications of micro-expressions. The validation accuracy of all evaluated architectures on the sleep/awake classification task is shown in Fig. \ref{fig:sleep_awakeGraph}.

For crying detection, Fig. \ref{fig:emotionsGraph} shows the convergence of the validation accuracy of all models during training and Fig. \ref{fig:cryingnormal} shows sample classification results based on the test set from the MobileNet. From Table \ref{tab:crying_model_performance}, MobileNet's performance was superior with a 97.69\% accuracy and a flawless AUC of 1.00, likely due to its ability to focus on salient features such as mouth strain and tear related facial distortions. Inception followed closely (94.44\% accuracy) but lagged in precision (0.94 vs. 0.98), likely due to rigid kernel sizes struggling with variable crying intensities. Both LeNet and ResNet152V2 plateaued at 90.74\% accuracy, reflecting their limitations in modelling the temporal progression of crying episodes. EfficientNetB0 again underperformed (78.24\% accuracy), reinforcing systemic flaws in its scaling strategy for neonatal contexts. 
Across both tasks, MobileNet demonstrated superior generalizability, ranking among the top two performers while maintaining compatibility with edge hardware. In contrast, Inception’s task-specific advantages—multiscale kernels for eye-state changes were offset by prohibitive memory demands. 
EfficientNetB0’s consistent underperformance highlighted its incompatibility with neonatal biomarker prioritization exposing flaws in its compound scaling strategy and shallow design for neonatal tasks. The model’s aggressive channel reduction likely discarded critical low level features (e.g., eyelid texture) necessary for sleep stage discrimination.
These findings underscore that clinical deployment demands architectures that balance accuracy with embedded constraints. 

\subsubsection{Comparative analysis benchmarking against the state-of-the-art models}

To contextualize the performance of the MobileNetV3 and InceptionV3 models, a comparative analysis was conducted against the current state of the art methods in infant monitoring.  The results, summarized in Table \ref{tab:comparative-performance}, demonstrate that our vision-based approach not only achieves highly competitive accuracy but does so through a fundamentally more practical and deployable paradigm.

\begin{table}[htbp]
\centering
\caption{Comparative performance of infant behavior classification models (current research vs. benchmarks }
\label{tab:comparative-performance}
\begin{tabular}{l l l c c c c c}
\toprule
\textbf{Task} & \textbf{Model} & \textbf{Modality} & \textbf{Acc (\%)} & \textbf{F1} & \textbf{AUC} & \textbf{Intrusive?} & \textbf{Edge-Deploy?} \\
\midrule
\multicolumn{8}{l}{\textbf{Our Research (Vision-Based)}} \\
Awake/Sleep & InceptionV3 & Image & 94.15 & 0.94 & 0.99 & \textbf{No} & No \\
Awake/Sleep & \textbf{MobileNetV3} & Image & 91.81 & 0.92 & 0.98 & \textbf{No} & \textbf{Yes} \\
Crying/Normal & InceptionV3 & Image & 94.44 & 0.94 & 0.99 & \textbf{No} & No \\
Crying/Normal & \textbf{MobileNetV3} & Image & \textbf{97.69} & \textbf{0.98} & \textbf{1.00} & \textbf{No} & \textbf{Yes} \\
\midrule
\multicolumn{8}{l}{\textbf{Benchmarks (Sensor-Based)}} \\
Pediatric Sleep & CNN+RNN \cite{JIMENEZGARCIA2024} & EEG/EOG/EMG & 84.1 & -- & -- & Yes & No \\
Pediatric Sleep & Ensemble DL \cite{phan2022} & PSG & 88.8 & 0.86 & -- & Yes & No \\
Pediatric Sleep & DNN \cite{Jeon2019} & EEG & 92.21 & 0.90 & -- & Yes & No \\
Awake/Sleep & DWT-CWT, ANN \cite{Irfan2025} & EEG & 90.37 & -- & -- & Yes & Potentially \\
Awake/Sleep & DWT-FAWT \cite{Irfan2024} & EEG (2-ch) & 87.56 & -- & -- & Yes & Potentially \\
\midrule
\multicolumn{8}{l}{\textbf{Benchmarks (Audio-Based)}} \\
Neonatal Cry & CNN-SVM \cite{Ashwini2021} & Audio & 88.89 & -- & 0.92 & No & No \\
Infant Cry & SE-ResNet-Trans \cite{Li2024} & Audio & 93.0 & 0.92 & -- & No & No \\
Infant Cry & MFCC + RF \cite{Hammoud2024} & Audio & 96.39 & -- & -- & No & No \\
Infant Cry & XGBoost \cite{chang2021} & Audio & 91.0 & -- & 0.9 & No & No \\
\bottomrule
\end{tabular}
\end{table}

%\subsubsection{Superior Performance and Practical Advantages of the Vision-Based Approach}
Our results establish that our non-contact, vision-based approach is highly competitive with the current state-of-the-art. For \textbf{awake/sleep detection}, our InceptionV3 model (94.15\% accuracy, 0.99 AUC) outperforms all cited sensor-based benchmarks \cite{JIMENEZGARCIA2024,phan2022} and performs on par with the most accurate EEG-based DNN \cite{Jeon2019}. Crucially, it achieves this without any intrusive sensors. Our edge-optimized MobileNetV3 model also performs strongly (91.81\% accuracy, 0.98 AUC), surpassing the CNN+RNN \cite{JIMENEZGARCIA2024} and ensemble \cite{phan2022} PSG benchmarks and closely matching the minimal-channel EEG results \cite{Irfan2024,Irfan2025}, all while being the only model in this comparison capable of real-time inference on a low-cost edge device.
For \textbf{crying detection}, our MobileNetV3 model's performance (97.69\% accuracy, 1.00 AUC) is exceptional, matching or exceeding the top audio-based benchmarks \cite{Hammoud2024,Li2024} that require high-quality microphones and are vulnerable to environmental noise. This demonstrates that visual cues are highly discriminative features for this task and can be captured effectively by an optimized vision model.

%\subsubsection{Redefining the State-of-the-Art: The Critical Axis of Deployability}
The most significant contribution of this work is revealed not just by the accuracy metrics, but by considering the dual axes of \textbf{performance} and \textbf{practical deployability}.
Existing high-accuracy benchmarks are largely confined to two categories: (1) intrusive, clinical-grade sensor systems \cite{JIMENEZGARCIA2024,phan2022,Irfan2024,Irfan2025,Jeon2019} and (2) non-intrusive but computationally complex audio models \cite{Ashwini2021,Li2024,Hammoud2024,chang2021} that require server-grade hardware and are susceptible to acoustic noise.
Our work creates a new category: \textbf{high-accuracy, non-intrusive, and edge-deployable}. As shown in Table~\ref{tab:comparative-performance}, our MobileNetV3 model is the only one that checks all three boxes. It delivers SOTA-tier performance for its specific tasks while running in real-time on a low-cost Raspberry Pi, processing data from a simple camera. This directly addresses the critical gap of the pervasive trade-off between clinical accuracy and practical deployability identified in the literature review.

%\subsubsection{Conclusion and Implications}
This comparative analysis confirms that our end-to-end optimized framework represents a paradigm shift. It moves the field from achieving high accuracy in controlled, resource-intensive environments to delivering high accuracy in practical, resource-constrained settings. The performance of our edge-deployable MobileNetV3 model proves that lightweight architectures, when properly optimized and applied to the right modality (vision), are not just compromises but can be superior, holistic solutions for real-world clinical deployment. This finding provides a compelling blueprint for the future development of accessible, non-invasive AI-powered healthcare tools for low-resource environments.

\subsection{Embedded performance and trade-offs}
The system's performance was rigorously evaluated on the Raspberry Pi platform (table~\ref{tab:tflite_benchmark})  to assess its suitability for real-time neonatal monitoring in resource-constrained environments. Our analysis reveals critical trade-offs between model complexity, inference latency, memory footprint, and overall system performance. These results underscore why model choice is critical for real-world deployment.

Benchmark results show both the runtime memory footprint (peak working set, MB), per-inference timing statistics (avg / std / min / max, in ms) and the stored TF-Lite binary size (MB).

\begin{table}[ht]
\centering
\caption{TF-Lite Inference Benchmarks on Raspberry Pi 5}
\label{tab:tflite_benchmark}
\begin{tabular}{lllllll}
\toprule
\textbf{Model} & \textbf{Memory (MB)} & \textbf{$T_{\text{mean}} \pm \text{std}$(ms)} & \textbf{$T_{\text{min}}$(ms)} & \textbf{$T_{\text{max}}$(ms)} & \textbf{Size(MB)} \\
\midrule
MobileNet   & 29.63  & $7.57 \pm 0.05$ & 7.47 & 7.65 & 2.59 \\
LeNet       & 32.41  & $9.82 \pm 0.09$ & 9.71 & 10.10 & 7.10 \\
ResNet      & 60.5  & $40.41 \pm 0.05$ & 40.33 & 40.60 & 10.70 \\
EfficientB0 & 160.92 & $61.08 \pm 1.86$ & 60.59 & 71.07 & 7.41 \\
InceptionV3 & 125.25 & $66.19 \pm 0.06$ & 66.06 & 66.40 & 23.00 \\
\bottomrule
\end{tabular}
\end{table}

\begin{figure}[ht]
  \centering
  \includegraphics[width=\linewidth]{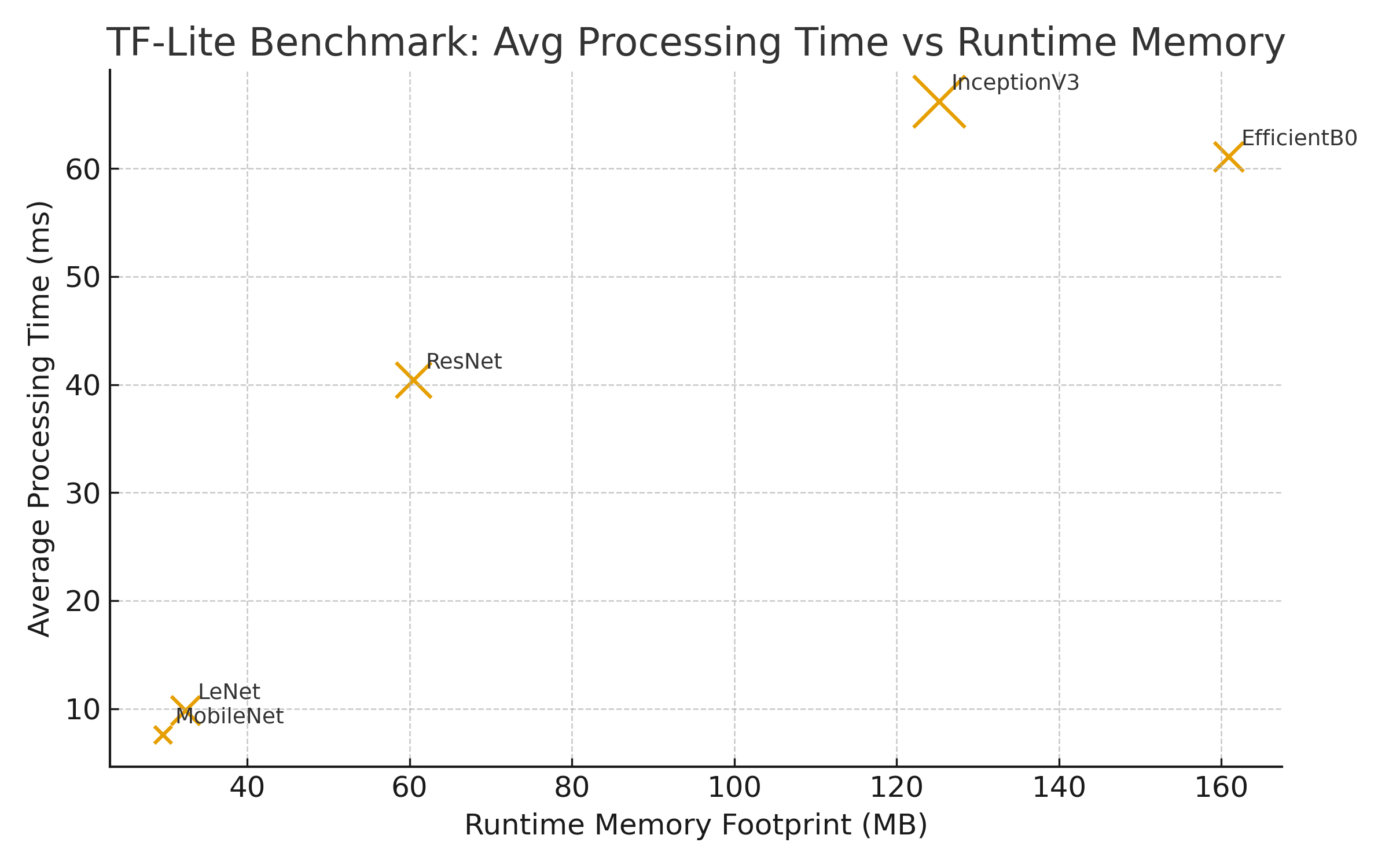}
  \caption{TF-Lite benchmark: average per-inference processing time versus runtime memory footprint. Marker size is proportional to the TF-Lite binary size.}
  \label{fig:tflite_scatter}
\end{figure}

\subsubsection{Interpretation and implications.}
The TF-Lite benchmark in Table~\ref{tab:tflite_benchmark} highlights several practical points relevant to on-device neonatal monitoring. First, there is a clear \textit{latency vs memory} trade-off. \textit{MobileNet} attains the lowest per-inference latency (7.57ms) and the smallest on-disk TF-Lite size (2.59 MB), making it well suited for continuous, low-latency operation on Raspberry Pi–class devices. \textit{LeNet} also provides fast inference ($\approx$ 9.8,ms) but may lack the representational capacity for clinically demanding tasks. In contrast, \textit{InceptionV3} and \textit{EfficientNetB0} impose substantially larger runtime memory footprints ($\approx$ 125–161MB) and longer processing times ($\approx$ 61–66ms), which limits their usability without hardware acceleration or aggressive compression.

Second, the results in the table underscores the important distinction of \textbf{runtime vs. storage memory} between stored model size and peak runtime working set. The fundamental trade-off between inference latency and runtime memory, as measured in our benchmarks, is visualized in Fig. \ref{fig:tflite_scatter}. EfficientNetB0, for example, shows a relatively small TF-Lite binary (7.41,MB) yet exhibits a large runtime footprint (160.92MB).  Highlighting how intermediate activations and temporary buffers dominate memory usage during inference.
 This behavior indicates that intermediate activation maps and temporary buffers , rather than only storage, drive peak RAM usage; thus, reporting both on disk and peak memory metrics is essential for realistic deployability assessments.
 
Third, the stability of the inference time matters for real-time guarantees. Most models show tightly clustered per-sample latencies (std $\approx $ 0.05–0.0 9ms), but EfficientNetB0 displays higher variability (std = 1.86ms and a pronounced max), suggesting occasional runtime spikes that could impair worst-case responsiveness. In safety-critical clinical applications the tail latency (max) can be more consequential than the mean; therefore, profiling and mitigating such spikes is required.

\subsubsection{Relation to system-level performance costs.}
These TF-Lite results should be read in the context of the pipeline complexity analysis (Table~\ref{tab:complexity_latency}). The per-frame costs are dominated by $\mathcal{O(HW)}$ operations (MediaPipe face/pose detection and ROI preprocessing). When media detection (10–60ms) is combined with per-face inference (5–20ms/face), the end-to-end latency for heavier models becomes infeasible for the 30FPS operation. Thus, a model that is compelling in isolation (e.g., high accuracy) may still be unsuitable for the full pipeline unless the model’s inference time and working set are reduced or offloaded to an accelerator. This reconciles the patterns observed in Tables \ref{tab:awake_model_performance},\ref{tab:crying_model_performance}, and  \ref{tab:tflite_benchmark}. MobileNet models are preferred for embedded deployment because they enable pipeline throughput, even if larger models offer marginally higher task accuracy.

\subsection{System-Level Performance Considerations}
\begin{table}[ht]
\centering
\caption{Complexity and latency analysis for core pipeline operations (720p, CPU)}
\label{tab:complexity_latency}
\begin{tabular}{p{0.35\linewidth} p{0.25\linewidth} p{0.3\linewidth}}
\toprule
\textbf{Operation} & \textbf{Complexity} & \textbf{Latency (720p, CPU)} \\
\midrule
Model Initialization & - & 100 ms - 1 s \\
Frame Reading & $\mathcal{O}(1)$ & 10-33 ms \\
Face Detection & $\mathcal{O}(HW)$ & 10-50 ms \\
Pose Detection & $\mathcal{O}(HW)$ & 20-60 ms \\
Per-face Inference & $\mathcal{O}(N)$ & 5-20 ms/face \\
Network Updates (Async) & $\mathcal{O}(1)$ & 50-200 ms \\
\bottomrule
\end{tabular}
\end{table}
The complete system performance must account for the entire processing pipeline, not just model inference. 
The complexity and latency analysis (Table \ref{tab:complexity_latency}) provides a concrete, operational perspective on why design choices in the algorithm \ref{alg:recognition_pipeline} pipeline materially affect real-time viability on Raspberry Pi class hardware.
Face and pose detection operations dominate processing time because of their $\mathcal{O}(HW)$ complexity.
At its core the analysis shows that per-frame work scales with image resolution (HW) and that face/pose detectors. Specifically MediaPipe in our implementation dominates the runtime. For 720p input the detectors and associated preprocessing together occupy the largest fraction of per-frame latency, and the marginal cost of handling multiple faces (N $\leq$ 5 in our target scenario) is small relative to the $\mathcal{O(HW)}$ term. Consequently, algorithmic complexity is effectively $\mathcal{O(HW)}$ under expected operating conditions. This asymptotic observation immediately explains the measured behavior reported in Section IV. the large models in Table III incur long worst-case latencies because they amplify the pixel-wise and region wise work already present in the detection stage.
The latency numbers in Table \ref{tab:complexity_latency} expose the practical bottlenecks. MediaPipe face and pose detection times (10–60 ms) combined with per-face inference (5–20 ms/face) can easily exceed the 33 ms budget required for 30 FPS operation. When pose detection and multi-face inference are enabled simultaneously, the cumulative latency frequently places the system in a sub-real-time regime unless hardware acceleration or resolution reduction is applied. Model initialization costs (100 ms – 1 s) and I/O overheads (frame read 10–33 ms) also contribute to initial stalls and tail latency but are amortized during continuous operation. Network updates to Firebase (50–200 ms) are correctly modelled as asynchronously. They affect throughput only in sofar as synchronous I/O is used, but they remain important for overall system responsiveness and for timely remote notifications.
\subsubsection{Practical implications}
First, resolution is a first-order lever. reducing input resolution from 720p to 480p or 360p scales down $\mathcal{O(HW)}$ cost and yields near linear reductions in detector time, at modest accuracy trade-offs for many behavioral cues. Second, temporal subsampling (frame skipping) and adaptive gating (lightweight motion detectors that trigger heavy pipelines only on activity) provide effective means to average down CPU load while preserving clinically relevant events that typically evolve over hundreds of milliseconds. Third, hardware acceleration (Edge TPU, Coral, or GPU) and model compression (quantization, pruning, knowledge distillation) are necessary to reconcile the representational needs of moderately complex models with hard resource constraints. These optimizations directly map to the trade-offs observed between Tables \ref{tab:awake_model_performance},\ref{tab:crying_model_performance} and \ref{tab:tflite_benchmark}. MobileNet-class networks are favorable from an efficiency perspective, but their representational limits explain why they underperform when the pipeline must simultaneously support multiple detection modes; larger networks preserve accuracy but fail latency and memory budgets.

\subsection{Future work}
 Future efforts will focus on closing the gap between laboratory validation and clinical utility. Immediate next steps include clinical trials to validate the system against manual scoring by healthcare professionals, collection of real-world NICU data with clinically diverse samples to improve model robustness against occlusions and lighting variations, and the exploration of hybrid architectures through knowledge distillation \cite{mugisha2025} to enhance lightweight models without compromising efficiency and evaluating the system in real-world NICU environments under ethical and regulatory oversight.

\section{Conclusions}
In this study, we present the development, optimization, and deployment of a vision-based embedded monitoring system for NICUs . Our work addresses the challenge of providing continuous neonatal care in low-resource settings. it goes beyond theoretical accuracy metrics and confronts the practical realities of implementing AI-driven solutions on affordable, resource-constrained hardware such as the raspberri pi and mobile devices.

The core finding of this research is that effective embedded deployment is achieved beyond simply selecting the most accurate model, but also consideration such as strategically balancing diagnostic performance with computational efficiency.

We identified and quantified the significant trade-offs between model architecture, inference latency, memory footprint, and system-level stability. We dscovered that While larger models such as the InceptionV3 achieve high standalone accuracy, their computational demands render them impractical for real-time use on a Raspberry Pi. Conversely, our optimized MobileNetV3 implementation emerged as the most viable foundation, achieving near-state-of-the-art accuracy (91.8\% for sleep/awake, 97.7\% for crying/normal) while maintaining a low latency profile and minimal memory footprint suitable for continuous operation.

An insight from our system-level evaluation was the "integrated performance paradox" , where models which excel in isolated tasks experienced significant accuracy drops when deployed within the full processing pipeline. This underscores the critical importance of evaluating models within the integrated complex, resource-contended environment of an end-to-end system and not just in isolation where factors such as face detection, preprocessing, and multitasking introduce new performance challenges.

The blueprint presented here integrates model quantization, hardware-aware optimization, and secure IoT communication. In addition, it provides a replicable framework for developing cost-effective medical informatics technologies. This work directly contributes to the goal of reducing global preterm mortality through accessible technological innovation. Through continued iterative design that prioritizes both algorithmic excellence and deployment feasibility, vision-based embedded systems hold significant promise for transforming neonatal care delivery in the world's most vulnerable communities.

%\bmhead{Supplementary information}: Not Applicable

\section*{Declarations}
\begin{itemize}
\item Funding: Not applicable
\item Conflict of interest/Competing interests: Not applicable
\item Ethics approval and consent to participate Not applicable : Not applicable
\item Consent for publication: All authors have given consent
\item Data availability : Not applicable
\item Materials availability : Not applicable
\item Code availability : Available after paper acceptance.
\item Author contribution : \textbf{Stanley Mugisha} did the Conceptualization, Methodology, Software, Validation, Formal Analysis, Investigation, Data Curation, Writing – Original Draft, Writing – Review and Editing, Visualization, Supervision, Project Administration. \textbf{Rashid Kisitu}: Methodology, Software, Validation, Investigation, Resources, Data Curation, Writing – Review and Editing. \textbf{Francis Komakech}: Conceptualization, Methodology, Software, Investigation, Writing – Review and Editing. \textbf{Excellence Favor}: Validation, Data Curation, and Formal Analysis,  Writing – Review and Editing.
\end{itemize}

%\noindent
%If any of the sections are not relevant to your manuscript, please include the heading and write `Not applicable' for that section. 

%\bigskip

%\bigskip\noindent
%Springer journals and proceedings: \url{https://www.springer.com/gp/editorial-policies}

%\bigskip\noindent
%Nature Portfolio journals: \url{https://www.nature.com/nature-research/editorial-policies}

\bigskip\noindent

\bibliography{sn-bibliography}% common bib file

\end{document}